\documentclass[letterpaper,twocolumn,10pt]{article}
\usepackage{hegroup}
\usepackage[numbers, compress]{natbib}
\usepackage{enumitem}
\usepackage[most]{tcolorbox}
\usepackage{mdframed}
\usepackage{multirow}
\usepackage{xspace}
\usepackage{hyperref}
\usepackage{tikz}
\usepackage{listings}
\usepackage{xcolor}
\usepackage{caption}
\usepackage{graphicx}
\usepackage{array}
\usepackage[flushmargin]{footmisc}
\usepackage{cleveref}
\usepackage{listings}

\definecolor{keywordcolor}{rgb}{0.2,0.2,1}
\definecolor{stringcolor}{rgb}{1,0.5,0}
\definecolor{commentcolor}{rgb}{0,0.6,0}
\definecolor{callcolor}{rgb}{0.8,0,0}
\newcommand{\GasAgent}{$\textit{GasAgent}$\xspace}

\definecolor{verylightgray}{rgb}{.97,.97,.97}
\lstdefinelanguage{Solidity}{
    keywords=[1]{
        anonymous, assembly, assert, balance, break, call, callcode, case, catch, class, constant, continue, 
        constructor, contract, debugger, default, delegatecall, delete, do, else, emit, event, experimental, 
        export, external, false, finally, for, function, gas, if, implements, import, in, indexed, instanceof, 
        interface, internal, is, length, library, log0, log1, log2, log3, log4, memory, modifier, new, payable, 
        pragma, private, protected, public, pure, push, require, return, returns, revert, selfdestruct, send, 
        solidity, storage, struct, suicide, super, switch, then, this, throw, transfer, true, try, typeof, using, 
        value, view, while, with, addmod, ecrecover, keccak256, mulmod, ripemd160, sha256, sha3
    },
    keywordstyle=[1]\color{blue}\bfseries,
    keywords=[2]{
        address, bool, byte, bytes, bytes1, bytes2, bytes3, bytes4, bytes5, bytes6, bytes7, bytes8, bytes9, 
        bytes10, bytes11, bytes12, bytes13, bytes14, bytes15, bytes16, bytes17, bytes18, bytes19, bytes20, 
        bytes21, bytes22, bytes23, bytes24, bytes25, bytes26, bytes27, bytes28, bytes29, bytes30, bytes31, 
        bytes32, enum, int, int8, int16, int24, int32, int40, int48, int56, int64, int72, int80, int88, int96, 
        int104, int112, int120, int128, int136, int144, int152, int160, int168, int176, int184, int192, int200, 
        int208, int216, int224, int232, int240, int248, int256, mapping, string, uint, uint8, uint16, uint24, 
        uint32, uint40, uint48, uint56, uint64, uint72, uint80, uint88, uint96, uint104, uint112, uint120, 
        uint128, uint136, uint144, uint152, uint160, uint168, uint176, uint184, uint192, uint200, uint208, 
        uint216, uint224, uint232, uint240, uint248, uint256, var, void, ether, finney, szabo, wei, days, 
        hours, minutes, seconds, weeks, years
    },
    keywordstyle=[2]\color{teal}\bfseries,
    keywords=[3]{
        block, blockhash, coinbase, difficulty, gaslimit, number, timestamp, msg, data, gas, sender, sig, 
        value, now, tx, gasprice, origin
    },
    keywordstyle=[3]\color{violet}\bfseries,
    identifierstyle=\color{black},
    sensitive=true,
    comment=[l]{//},
    morecomment=[s]{/*}{*/},
    commentstyle=\color{gray}\ttfamily,
    stringstyle=\color{red}\ttfamily,
    morestring=[b]',
    morestring=[b]"
}

\newcommand{\mypara}[1]{\smallskip\noindent{\bf {#1}.}\xspace}

\date{}

\author{
\textbf{Jingyi Zheng}\thanks{Contributed equally.}\enspace\enspace 
\textbf{Zifan Peng}\textsuperscript{\textcolor{blue!60!green}{$\ast$}}\enspace\enspace
\textbf{Yule Liu}\enspace\enspace
\textbf{Junfeng Wang}\enspace\enspace\textbf{Yifan Liao}\enspace\enspace
\textbf{Wenhan Dong}\enspace\enspace
\textbf{Xinlei He}\thanks{Corresponding author (\href{mailto:xinleihe@hkust-gz.edu.cn}{xinleihe@hkust-gz.edu.cn}).}
\\ 
The Hong Kong University of Science and Technology (Guangzhou)
}

\begin{document}

\title{\GasAgent: A Multi-Agent Framework for Automated \\ Gas Optimization in Smart Contracts}

\maketitle

\begin{abstract}
Smart contracts are trustworthy, immutable, and automatically executed programs on the blockchain.
Their execution requires the Gas mechanism to ensure efficiency and fairness.
However, due to non-optimal coding practices, many contracts contain Gas waste patterns that need to be optimized.
Existing solutions mostly rely on manual discovery, which is inefficient, costly to maintain, and difficult to scale.
Recent research uses large language models (LLMs) to explore new Gas waste patterns.
However, it struggles to remain compatible with existing patterns, often produces redundant patterns, and requires manual validation/rewriting.
To address this gap, we present \textbf{\GasAgent}, the first multi-agent system for smart contract Gas optimization that combines compatibility with existing patterns and automated discovery/validation of new patterns, enabling end-to-end optimization.
\GasAgent consists of four specialized agents—\textit{Seeker}, \textit{Innovator}, \textit{Executor}, and \textit{Manager}—that collaborate in a closed loop to identify, validate, and apply Gas-saving improvements.
Experiments on 100 verified real-world contracts demonstrate that \GasAgent successfully optimizes 82 contracts, achieving an average deployment Gas savings of 9.97\%.
In addition, our evaluation confirms its compatibility with existing tools and validates the effectiveness of each module through ablation studies.
To assess broader usability, we further evaluate 500 contracts generated by five representative LLMs across 10 categories and find that \GasAgent optimizes 79.8\% of them, with deployment Gas savings ranging from 4.79\% to 13.93\%, showing its usability as the optimization layer for LLM-assisted smart contract development.
\end{abstract}

\section{Introduction}

\begin{figure*}
  \centering
  \includegraphics[width=\linewidth]{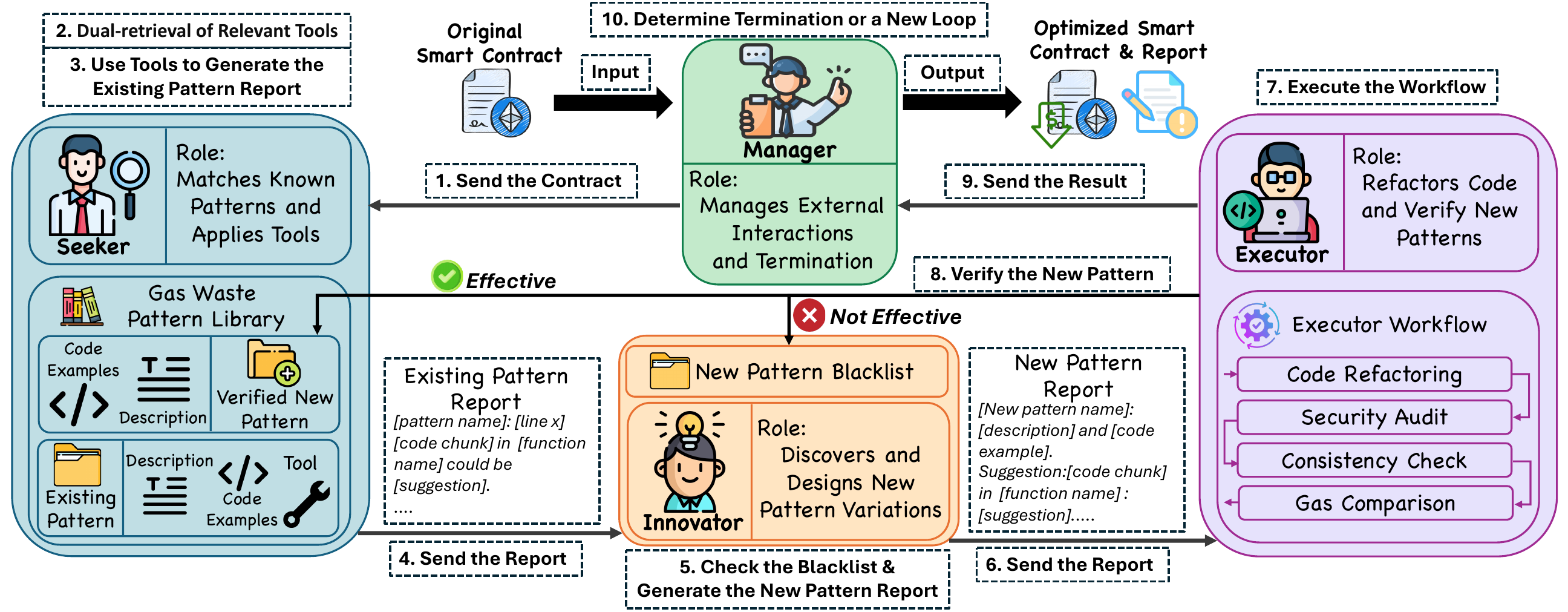}
  \caption{The workflow of \GasAgent, where the Seeker, Innovator, Executor, and Manager form a closed loop for automated Gas optimization via pattern matching, new discovery, and verification.}
  \label{fig:gasagent-workflow}
\end{figure*}

Blockchain enables trustless collaboration by decentralizing computation and storage.
The introduction of smart contracts, which are programs that automatically execute predefined logic on the nodes~\cite{zou2019smart, von2016blockchain}, has significantly expanded blockchain applications.
These applications include decentralized finance (DeFi), decentralized autonomous organizations (DAOs), and other scenarios requiring transparent and tamper-proof automation.
To manage limited resources, Ethereum introduced the Gas mechanism, which quantifies the cost of deploying and executing smart contracts in units called \emph{wei}, each value at $10^{-18}$ Ether.
However, many developers lack familiarity with the Gas pricing model, and conventional coding practices frequently result in inefficiencies.
Thus, deployed contracts often contain redundant logic that wastes Gas~\cite{chen2020gaschecker, liu2020understanding}.
As smart contract adoption grows, improving Gas efficiency has become an important research topic~\cite{kaur2022gas,jiang2024unearthing,chen2020gaschecker}.

Most existing approaches to Gas optimization depend on manually defined patterns and human inspection, which limits their scalability and adaptability to new inefficiencies~\cite{albert2020gasol, kaur2022gas, zhao2022gasaver}.
Recent research~\cite{jiang2024unearthing} has explored the use of large language models (LLMs) to discover potential Gas waste patterns, which represents an important step toward automation.
However, many of the generated patterns may overlap with existing ones or contain hallucinations, requiring users to manually verify their validity and determine appropriate code refactors, which still demands significant expertise.
More importantly, existing works focus solely on identifying inefficiencies, without supporting end-to-end automation of the entire optimization workflow.
Ideally, such a workflow would include not only pattern discovery but also automated pattern judgement, code refactoring, and various checking, aligning with the rising desire for fully automated development paradigms such as vibe-coding~\cite{maes2025gotchas}.
Yet in practice, a single LLM often struggles with multi-step tasks that require decomposing problems, verifying correctness, and enforcing semantic consistency~\cite{hong2023metagpt, chen2021evaluating}.
These limitations highlight a critical gap: \textbf{How can we automatically and effectively discover comprehensive patterns while also enabling automated verification and code refactoring?}

Smart contract Gas optimization naturally decomposes into sub-tasks such as identifying inefficiencies, proposing optimizations, and verifying effectiveness.
This modularity makes it well-suited for multi-agent systems, where specialized agents can collaborate, debate, and verify each other’s outputs—forming a natural and scalable foundation for an end-to-end optimization pipeline~\cite{guo2024large, du2023improving}.
In this paper, we propose \textbf{\GasAgent}, a multi-agent smart contract Gas optimization system that is self-updating and fully compatible with existing Gas waste patterns.
\GasAgent addresses three key challenges in applying LLMs to Gas optimization:
(1) ensuring that the model reliably covers all existing Gas waste patterns; 
(2) preventing high redundancy between new and existing patterns;
and (3) automatically verifying new patterns and applying them for code refactoring.
\GasAgent is composed of four specialized agents: The \textbf{\textit{Seeker}}, which retrieves existing Gas waste patterns from a continuously evolving pattern library;
The \textbf{\textit{Innovator}} leverages the LLM to propose new optimization patterns beyond the current library, ensuring that the system adapts to novel contract structures;
(3) The \textbf{\textit{Executor}} applies and validates the suggested changes through code refactoring, security audits, consistency checks, and Gas savings measurements.
(4) The \textbf{\textit{Manager}} handles external interactions, determines when to terminate the loop, and generates reports for human review.
We also design a continuously evolving pattern library that enables \GasAgent to maintain self-updating capabilities by integrating new Gas waste patterns as they are discovered and verified.
This architecture draws inspiration from how experienced contract developers identify costly code fragments, reason about their impact, and iteratively test optimizations to save Gas while maintaining functional correctness.

To verify the effectiveness of \GasAgent, we collected 100 real-world contracts from Etherscan.
Our results show that \GasAgent achieves an average deployment Gas saving of 9.97\%, successfully optimizing 82\% of them.
To assess compatibility with prior work, \GasAgent recalls 92.5\% of 557 ground-truth pattern instances defined by 24 existing tools, while reducing detection calls by 28.2\% through efficient retrieval strategies.
The ablation study shows that the full \GasAgent system outperforms all variants, optimizing 82 contracts (saving 9.97\%) compared to 71 contracts (saving 5.93\%) for direct LLM optimizing, confirming the necessity of both the Seeker and Innovator modules.
To evaluate its usability in LLM-assisted development, we further evaluate whether \GasAgent can optimize LLM-generated contracts.
Applied to 500 contracts generated by 5 representative LLMs, \GasAgent successfully optimizes 79.8\%, with model-wise average deployment savings ranging from 4.79\% to 13.93\%, demonstrating its usability as a reliable optimization layer for LLM-assisted smart contract development.

In summary, this paper makes the following contributions:

\begin{itemize}[leftmargin=*]
    \item We present \GasAgent, the first multi-agent framework for smart contract Gas optimization, combining compatibility with existing Gas waste patterns and discovery of new ones to enable end-to-end optimization automatically.
    
    \item We conduct comprehensive evaluations on verified real-world contracts to demonstrate \textit{GasAgent}’s effectiveness.
    Our experiments include ablation analysis to validate each module’s necessity and compatibility testing to confirm the reuse of existing gas waste patterns.

    \item We further show that \GasAgent effectively handles LLM-generated contracts, demonstrating its utility as a plug-and-play optimization layer in LLM-assisted smart contract development to reduce Gas waste across broader scenarios.
\end{itemize}

\section{Background}

\subsection{LLMs and Multi-Agent Systems}

LLMs represent a breakthrough in artificial intelligence, achieving impressive performance in natural language understanding and generation through Transformer-based architectures and large-scale pretraining~\cite{vaswani2017attention,hou2024large,yang2024harnessing}.
Notable models include BERT~\cite{devlin2019bert}, which introduced bidirectional context encoding; GPT series~\cite{radford2019language}, which leveraged autoregressive generation; and T5~\cite{raffel2020exploring}, which unified NLP tasks under a text-to-text framework.
LLMs have also been extended to programming tasks, with models like CodeBERT~\cite{feng2020codebert}, CodeT5~\cite{wang2021codet5}, and Codex~\cite{chen2021evaluating} excelling in code understanding and generation.
Recent proprietary models such as GPT-4~\cite{achiam2023gpt} and Gemini~\cite{team2023gemini} further push the boundaries with better instruction following and domain adaptation capabilities.
Agents built on top of LLMs can perceive environments, reason, and act toward specific goals, making LLMs powerful backbones for intelligent behavior~\cite{russell2016artificial, wooldridge1995intelligent}.
In single-agent settings, LLMs can decompose tasks~\cite{khot2022decomposed}, invoke external tools~\cite{li2023api,ruan2023tptu}, and utilize memory~\cite{dong2022survey} to complete complex workflows~\cite{sumers2023cognitive}.
However, they often struggle with scalability, consistency, and modular reasoning in multi-step tasks.
To overcome these limitations, LLM-based multi-agent systems (LLM-MA) employ multiple interacting agents with diverse roles~\cite{guo2024large,mandi2024roco}, which communicate~\cite{liu2023dynamic}, collaborate~\cite{du2023improving}, and cross-check each other to enhance performance and adaptability.
LLM-MA have shown great promise in software engineering~\cite{hong2023metagpt,qian2023communicative}, robotics~\cite{zhang2023building}, policy and society simulation~\cite{xiao2023simulating,park2023generative}, and game environments~\cite{xu2023language}.
Compared to single-agent, LLM-MA can distribute responsibilities, enhance reasoning, and better scale to real-world multi-step problems.

\subsection{Smart Contracts and Gas Optimizations}

Blockchain is a decentralized ledger technology that lets multiple untrusted nodes keep a consistent and immutable record of transactions without an authority, using cryptography and distributed consensus~\cite{von2016blockchain}.
Modern blockchain systems support smart contracts, which are computer programs that automatically execute predefined logic on every participating node in the network~\cite{zou2019smart}.
This trustless automation enables smart contracts to power applications in decentralized finance, digital payments, supply chains, and other collaborative scenarios where trusted third parties are replaced by code~\cite{zheng2020overview}.
However, running smart contracts consumes computing and storage resources on every node in the network.
Due to the fully replicated execution model, blockchain inherently has limited transaction throughput compared to centralized systems~\cite{wood2014ethereum}.
Although techniques such as sharding~\cite{xu2023two} have been explored to improve throughput, the fundamental resource limitation of storing and executing smart contracts remains.
Blockchain systems like Ethereum use a Gas system to prevent resource abuse~\cite{chen2017adaptive}.
Gas is used to measure how much computing power and storage are required to run a transaction or a smart contract~\cite{albert2020gasol}.
When a developer deploys a contract or a user calls a contract, they must pay Gas fees based on the contract's size and the complexity of its operations.
This makes sure that people pay for the resources they use and discourages wasteful or malicious actions, such as attacks that try to overload the network~\cite{liu2020understanding}.
The Gas mechanism ensures that system resources are used properly, but many contracts still include redundant or costly code because developers often lack the tools or experience to write optimized code, which can make them cost much more than necessary~\cite{chen2020gaschecker}.
Therefore, improving Gas efficiency is important to cut additional costs, save network resources, and help blockchains handle more useful work.

\section{Methodology}

\GasAgent is a multi-agent framework designed to automate smart contract Gas optimization by integrating known pattern detection, new pattern discovery, code refactoring, and automatic verification.
In this section, we first describe the design motivation and key challenges that shape \textit{GasAgent}’s architecture (\Cref{sec:design-motivation}).
We then present the system overview (\Cref{sec:system-overview}) and detail the four specialized agent roles: Seeker (\Cref{sec:seeker}), Innovator (\Cref{sec:innovator}), Executor (\Cref{sec:executor}), and Manager (\Cref{sec:Manager}).

\subsection{Design Motivation}
\label{sec:design-motivation}

Our goal is to develop an automated system that optimizes smart contract Gas usage with minimal human involvement while ensuring functional correctness and security.
To achieve this, we identify several essential capabilities that the system needs to support.
These capabilities directly motivate the multi-agent architecture and workflow adopted by \GasAgent, where each agent is designed to fulfill a specific role in the end-to-end optimization pipeline.

\textbf{(1) Integration with Known Existing Gas Waste Patterns.} 
Existing Gas waste patterns are code structures confirmed by auditors or researchers to cause unnecessary Gas consumption.
Pre-trained LLMs are trained on general text and code, with knowledge fixed at the time of training; thus, they typically do not include Gas waste patterns discovered later by auditors or researchers.
A recent study~\cite{jiang2024unearthing} shows that pre-trained LLMs miss known Gas waste patterns in practice, suggesting that even patterns discovered before pre-training may not be fully captured or recalled.
Although fine-tuning is a potential solution, it demands high-quality data and significant computational overhead, and currently lacks publicly available datasets for Gas optimization.

To achieve this, \GasAgent introduces an agent called Seeker, which leverages a dual retrieval mechanism to align contracts with an updatable external pattern library, ensuring integration of existing patterns without retraining the base LLM.
By tuning retrieval parameters, Seeker allows flexible trade-offs between recall and computational consumption, adapting to different optimization scenarios.

\textbf{(2) Discovering Novel Optimizations While Remaining Grounded in Existing Knowledge.} 
An effective Gas optimization system should not only incorporate known existing patterns but also explore novel patterns that go beyond known templates.
As smart contract practices evolve, new inefficiencies may appear in forms that deviate from prior cases.
Relying solely on fixed patterns risks overlooking these emerging opportunities.
At the same time, unconstrained generation may lead to unrealistic or irrelevant suggestions.
Therefore, the system must strike a careful balance that grounds the LLM in existing patterns while guiding it to propose innovative yet plausible optimizations.

To balance this, \GasAgent introduces another agent called Innovator, which builds on top of the Seeker’s results and guides the LLM to suggest new or refined patterns with the confirmed ones as context.
This design helps keep the system conservative where needed while still enabling low-cost exploration of new patterns.

\textbf{(3) Validating the Safety and Effectiveness of Proposed Pattern.} 
When the LLM proposes new Gas waste patterns, its output may include hallucinations, such as ideas that seem reasonable in text but fail in practice or are completely incorrect.
Blindly trusting the raw output for further processing is risky, as such changes could break contract logic or introduce new security issues.

To address this, \GasAgent includes an agent called Executor, which systematically rewrites the contract based on suggestions from the Seeker and Innovator and runs structured checks that contain security, functional correctness, and actual Gas savings.
This process ensures that only effective and safe optimizations are retained.

\subsection{System Overview}
\label{sec:system-overview}

To address the above challenges, \GasAgent adopts a modular multi-agent architecture combining existing pattern matching, new pattern discovery, and systematic verification in a closed loop.
This is achieved by four specialized agents that divide and refine the tasks:  detecting existing patterns, discovering new patterns, verifying outputs, and managing the workflow.
Concretely, \GasAgent has these specialized roles:

\begin{itemize}[leftmargin=*]

    \item \textbf{Seeker}: The Seeker is responsible for identifying known Gas inefficiencies by matching the target smart contract against patterns stored in the existing pattern library.
    For each match, it generates detailed reports that provide evidence of relevant code locations and suggest improvements.
    The Seeker aims to ensure that all Gas wastes covered by existing patterns are identified.
    It works like an experienced inspector who never misses obvious mistakes.

    \item \textbf{Innovator}: The Innovator focuses on discovering new or improved Gas-saving patterns that are not yet included in the existing pattern library.
    It proposes possible new patterns or tweaks old ones, checks them against a blacklist to filter out invalid ideas, and reports valid candidates for further testing and validation.
    This ensures that \textit{GasAgent}’s pattern library can evolve and adapt to new coding styles or emerging inefficiencies.
    It acts like a creative developer who always brings fresh ideas.
    
    \item \textbf{Executor}: The Executor applies code refactoring based on the reports from the Seeker and Innovator, respectively.
    Then it verifies every change by doing a security audit, functional consistency check, and Gas cost comparisons to make sure the suggestions really work as intended without introducing new risks.
    It works like a reliable craftsman who checks every detail to make sure all ideas are both safe and effective.

    \item \textbf{Manager}: The Manager handles external interactions and decides when to terminate the optimization loop.
    It determines the final solution and generates reports for human review.
    It works like a team leader who ensures smooth communication with external stakeholders and the delivery of optimized results.
    
\end{itemize}

\begin{table}[t]
\centering
\caption{Schema of a pattern entry in the Gas Waste Pattern Library, contains fields for top-level metadata and detailed example objects.}
\label{tab:pattern-schema}
\resizebox{\linewidth}{!}{
\begin{tabular}{>{\centering\arraybackslash}m{1cm} >{\raggedright\ttfamily\arraybackslash}m{3.2cm} m{6cm}}
\toprule
\textbf{} & \textbf{Field} & \textbf{Description} \\
\midrule
\multirow{6}{*}{\rotatebox{90}{\textit{Top-level Fields}}}
  & name & Unique identifier for the pattern. \\
  & description & Explanation of the Gas waste scenario. \\
  & summary & Short statement of the key idea. \\
  & tags & Keywords for quick search. \\
  & applicableScenarios & Contexts where this pattern applies. \\
  & examples & List of example objects (fields below). \\
\midrule
\multirow{10}{*}{\rotatebox{90}{\textit{Fields in Each Example}}}
  & id & Unique ID for the example. \\
  & title & Short title describing the use case. \\
  & description & Explains what the example shows. \\
  & codeBefore & Original Solidity code showing the inefficient practice. \\
  & codeAfter & Optimized version showing the recommended improvement. \\
  & codeIssueTags & Tags for related code features. \\
  & codeImprovements & List of concrete improvements or Gas savings achieved. \\
\bottomrule
\end{tabular}
}
\end{table}

\mypara{Workflow of \GasAgent}
\Cref{fig:gasagent-workflow} shows the workflow of \GasAgent.
The process begins with the Seeker, which analyzes the original smart contract and performs dual retrieval over the Gas Waste Pattern Library to generate an Existing Pattern Report containing matched patterns and recommended fixes.
The Innovator then takes this report as context to design new pattern variations beyond the existing library, filtering out invalid or redundant suggestions using a maintained blacklist.
Next, the Executor integrates both reports, refactors the contract accordingly, and runs a structured verification pipeline—including code rewriting, security auditing, consistency checking, and Gas usage comparison—to ensure all changes are safe and effective.
Patterns that fail validation are blacklisted, while successful ones are incorporated back into the verified pattern library.
Throughout this loop, the Manager manages external inputs and outputs, collects results, generates reports, and determines whether to terminate or initiate a new optimization loop.

\subsection{Seeker}
\label{sec:seeker}

\begin{figure}[t]
  \centering
  \includegraphics[width=\linewidth]{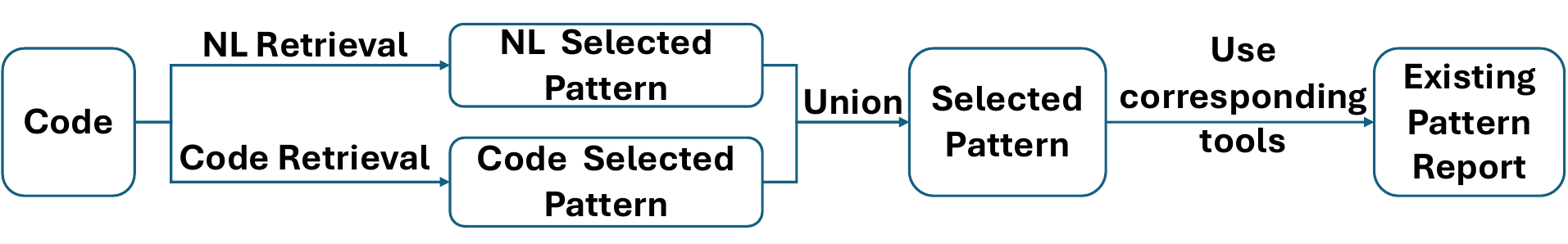}
  \caption{Workflow of the Seeker. Patterns are retrieved from the Gas Waste Pattern Library using both Natural Language (NL) and code similarity.}
  \label{fig:seeker-workflow}
\end{figure}

The Seeker is designed to overcome the limitations of pre-trained LLMs by ensuring that all known Gas waste patterns are reliably detected.
The workflow is illustrated in \Cref{fig:seeker-workflow}.
At the outset, the Seeker constructs a Pattern Library based on prior research.
This library is structured as a directory of JSON files, where each file represents a verified pattern and includes associated metadata and concrete code examples.
\Cref{tab:pattern-schema} summarizes the schema of each pattern entry, detailing both top-level fields and the internal structure of example entries.
The Pattern Library ensures that \GasAgent remains aligned with expert-curated optimization knowledge.

During contract optimization, the Seeker employs two complementary retrieval strategies, i.e., code-based and Natural Language-based retrieval, to identify relevant Gas waste patterns.
In code-based retrieval, the Seeker encodes the input contract along with corresponding example code snippets from the pattern library into embeddings, then computes code similarity using cosine similarity.
Patterns whose example similarity exceeds a predefined threshold are selected.
In parallel, the Seeker performs Natural Language-based retrieval by generating a prompt that combines the contract’s source code with natural language descriptions of all known patterns.
This prompt is sent to the LLM, which returns a list of matched pattern IDs based on semantic understanding.
Finally, the two sets of selected patterns are merged for further analysis, which is further compiled into a structured Existing Pattern Report, forming a foundation for optimization.

\begin{tcolorbox}[title=Seeker Prompt in Natural Language-based Retrieval, breakable, colback=gray!5!white, colframe=black!75!white, fonttitle=\bfseries]
\textbf{System Prompt:} 

You are a smart contract analysis expert.
Please analyze the given contract code and select relevant patterns from the provided optimization pattern list (no limit on quantity) to optimize the Gas Fee.
Only return pattern IDs, separated by commas, e.g., \texttt{repeated\_computation, state\_variable\_refactoring}.

\textbf{User Prompt:}  

Please analyze the following smart contract and select the patterns that need optimization:

\texttt{\small [Contract Source Code Here]}

Below are the provided optimization patterns:  

\texttt{\small [Pattern Descriptions Here]}

Please only return pattern IDs, separated by commas.
\end{tcolorbox}

Next, the Seeker invokes the corresponding Python tool to apply the retrieved patterns to the original code and run fine-grained analysis.
The tool validates whether the applied patterns and the extracted structural information are accurate.
The results are then compiled into the Existing Pattern Report, which will be sent to the Innovator for novel pattern discovery and to the Manager for system-level oversight.

\subsection{Innovator}
\label{sec:innovator}

The Innovator is designed to address the limitation that relying solely on known patterns is insufficient for achieving better Gas optimization, especially as smart contract development styles continue to evolve.
Building upon the output from the Seeker, the Innovator uses confirmed pattern matches as contextual grounding for LLMs, enabling them to identify novel Gas inefficiencies or variations of existing patterns.
Specifically, the LLM is explicitly prompted to avoid duplicating existing suggestions and to generate actionable ideas that directly reference specific portions of the input code.

\begin{tcolorbox}[title={Innovator Prompt}, breakable, colback=gray!5!white, colframe=black!75!white, fonttitle=\bfseries]

\textbf{System Prompt:} 

Please try to summarize a new Gas optimization pattern based on the current contract code and existing suggestions, with the main goal of reducing Gas fees.
\textit{Note:} The suggestion should be different from the existing ones.
Do not repeat existing suggestions.
Please specify which parts of the current contract code match the new pattern.
The generated patterns must be reasonable; if none are found, it's okay not to generate any.
And if there are multiple patterns, please just generate the most important one.

\textbf{User Prompt:}  

Current contract code:

\texttt{\small [Contract Source Code Here]}

Existing suggestions:

\texttt{\small [Seeker’s Suggestions Here]}

\end{tcolorbox}

For each proposed pattern, the Innovator first checks the New Pattern Blacklist (the list of previously proposed but invalidated patterns) to ensure that the new propositions are novel.
Once validated, the proposed pattern is compiled into a New Pattern Report, which includes a proposed name, a description, the relevant code segments, and an explanation of how the pattern could reduce Gas consumption.
This report is then forwarded to the Executor for structured verification.

\subsection{Executor}
\label{sec:executor}

The Executor acts as the safeguard and checking point in the \GasAgent workflow, ensuring that any contract optimizations proposed by the Seeker and Innovator are safe and effective.
The process consists of code refactoring, security audit, consistency check, and Gas cost comparison.
The workflow begins by taking the original smart contract and refactoring it using the suggestions in the Existing Pattern Report generated by the Seeker.
This produces the first optimized code version, which then undergoes a full validation pipeline as shown in~\Cref{fig:gasagent-workflow}.

Next, the Executor conducts a Security Audit using Slither~\cite{feist2019slither} to detect potential vulnerabilities introduced during rewriting, followed by automatic generation of a differential testing suite to perform a Consistency Check between the optimized and original contract versions.
This suite covers normal unit tests, boundary-value inputs, and fuzzing to exercise edge cases, covering the key aspects summarized in function scope, input diversity, randomized fuzzing, deployment init, behavior equivalence, and structural tolerance.
If some step fails—Security Audit, Consistency Check—the refactored code of the Seeker version is discarded, and the original contract is retained.
Finally, the Executor measures and compares the Gas cost of the refactored contract to the original to confirm the effectiveness of the optimizations from the Seeker.
Both the Security Audit and the Consistency Check are designed to be fully modular and extensible, so they can be enhanced with more advanced analyzers, alternative checkers, or more comprehensive testing presets as needed.\footnote{However, extending these components is not the focus of this work; in this design, we only provide the basic implementation necessary to validate the optimization.}

If the optimization from the Seeker passes all checks, the Executor applies the second stage: refactoring the code using the New Pattern Report from the Innovator.
This stage generates the second optimized code version on top of the first version.
The second version undergoes the same comprehensive verification: Security Audit, Consistency Check, and Gas Comparison.
The new version must pass all checks and demonstrate strictly lower Gas costs than the first optimized code version.
If so, the new pattern proposed by the Innovator is deemed valid and added to the Gas waste pattern library, though its corresponding Python tool still needs to be manually implemented.
If the second version passes all the checks, the new pattern proposed is blacklisted by the Innovator to avoid redundant future exploration.

\subsection{Manager}
\label{sec:Manager}

The Manager serves as the external-facing controller responsible for overseeing the progress of the \GasAgent workflow.
It interacts with users or calling systems, collects the outputs from internal agents, and determines when the optimization process should terminate.

A central responsibility of the Manager is deciding when further optimization is no longer beneficial.
After each iteration, it reviews the results produced by the Seeker, Innovator, and Executor—including pattern matches, new suggestions, validation outcomes, and measured gas savings.
If an optimization is verified effective, the Manager allows another round of exploration to continue; otherwise, it halts the loop and finalizes the last validated contract.
Throughout the process, the Manager compiles structured reports summarizing key actions and decisions across agents, making the full optimization trace transparent and interpretable to end users.
It works like a client-facing manager who ensures that the team’s collective effort results in a coherent and justifiable outcome before returning it to the outside world.

\section{Evaluation}

\subsection{Research Questions (RQs)}

Since \GasAgent is designed as a fully automated framework that combines comprehensive existing pattern-based Gas optimization, novel pattern discovery, and validation through a multi-agent structure, we conduct a series of experiments to evaluate its practical effectiveness, compatibility, design rationality, and usability.
Our experiments aim to answer the following research questions:

\begin{enumerate}[label=\textbf{RQ\arabic*:}, left=0pt, itemindent=!, align=left]

    \item \textbf{Effectiveness - Is \GasAgent effective in reducing Gas fees?}
    This question explores whether the \GasAgent can deliver measurable Gas savings.
    We measure effectiveness by evaluating how much \GasAgent can optimize real-world smart contracts that are already deployed and verified on-chain.
    This question also concerns whether the new Gas waste patterns proposed by \GasAgent are reasonable.

    \item \textbf{Pattern Incorporation – Can \GasAgent comprehensively integrate and reuse existing Gas waste patterns proposed in prior work?}
    This question explores whether \GasAgent can faithfully incorporate existing patterns proposed by prior works into its internal library and leverage them via the dual retrieval mechanism.

    \item \textbf{Design Rationality - Are the roles of individual agents in \GasAgent well-motivated and indispensable?}
    This question evaluates whether agents like the Seeker and the Innovator serve distinct, necessary functions within the framework or whether some agents could be omitted.

    \item \textbf{Broader Usability - Can \GasAgent act as an effective optimization layer for LLM-assisted smart contract development?} 
    This question investigates whether \GasAgent can be reliably applied as an optimization layer to detect and optimize redundant Gas usage in smart contracts generated by LLMs.
    In addition, it examines whether outputs from different LLMs benefit differently from \GasAgent optimization, highlighting variations in their remaining inefficiencies.

\end{enumerate}

\subsection{Experimental Setting}
\label{sec:setting}

Our \GasAgent is implemented in Python and orchestrated using LangGraph~\cite{langgraph} to manage all agents.
All LLM-driven tasks use \texttt{GPT-4o-2024-11-20}~\cite{gpt4o} via the OpenAI API.
For code similarity retrieval in the Seeker, we apply the \texttt{jina-v2}~\cite{gunther2023jina} embedding model and select any pattern whose code example cosine similarity with the input code exceeds a fixed threshold of 0.7.
All contracts are compiled with solc version 0.8.20.
Security checks use Slither~\cite{feist2019slither} with the same compiler version to ensure compatibility.
Deployment Gas cost measurements are cross-validated using both Ganache~\cite{ganache} and Hardhat~\cite{hardhat} local test networks to guarantee consistent Gas estimates across different Ethereum Virtual Machine(EVM) backends.
For consistency checks, \GasAgent automatically generates unit tests, boundary-value tests, and fuzzing tests using Foundry~\cite{foundry}, with a maximum of 5 parameter combinations and 100 fuzz runs per function by default.
The Gas waste pattern library includes 24 patterns that have been explicitly identified in prior research~\cite{jiang2024unearthing,kaur2022gas,zhao2022gasaver,di2022profiling,nguyen2022gassaver,marchesi2020design}.
We employed four PhD students in computer science to read these papers and implement each pattern as a reusable Python module.
which can automatically detect the corresponding inefficiency in any given Solidity contract.
Our metric focuses on deployment Gas, which often exhibits similar trends to message-call Gas consumption.
While we currently use deployment Gas to evaluate new pattern effectiveness via the executor, the same pipeline supports message-call gas as a replacement if needed.

\begin{figure}[t]
  \centering
  \includegraphics[width=0.97\linewidth]{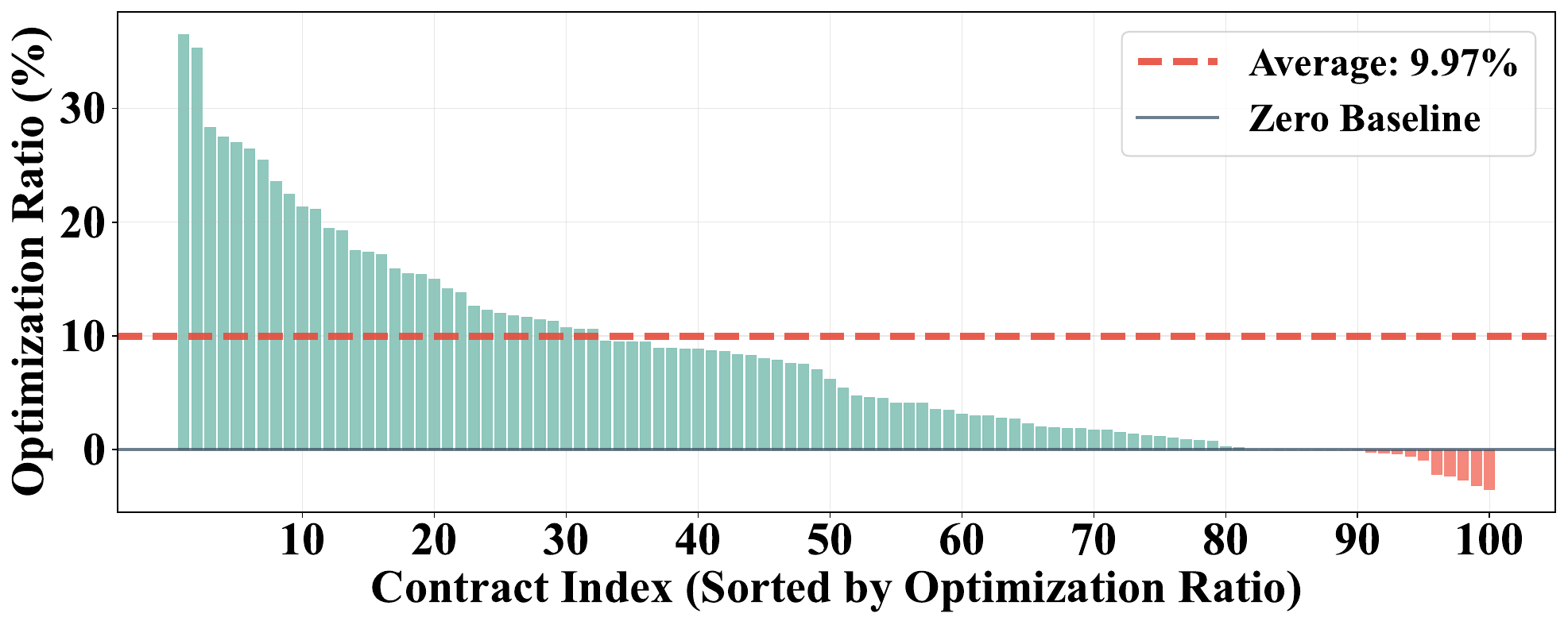}
  \caption{Gas optimization ratio distribution across 100 real-world smart contracts. The average saving is 9.97\% compared to the original versions.}
  \label{fig:real-gas-distribution}
\end{figure}

\begin{figure}[t]
  \centering
  \includegraphics[width=\linewidth]{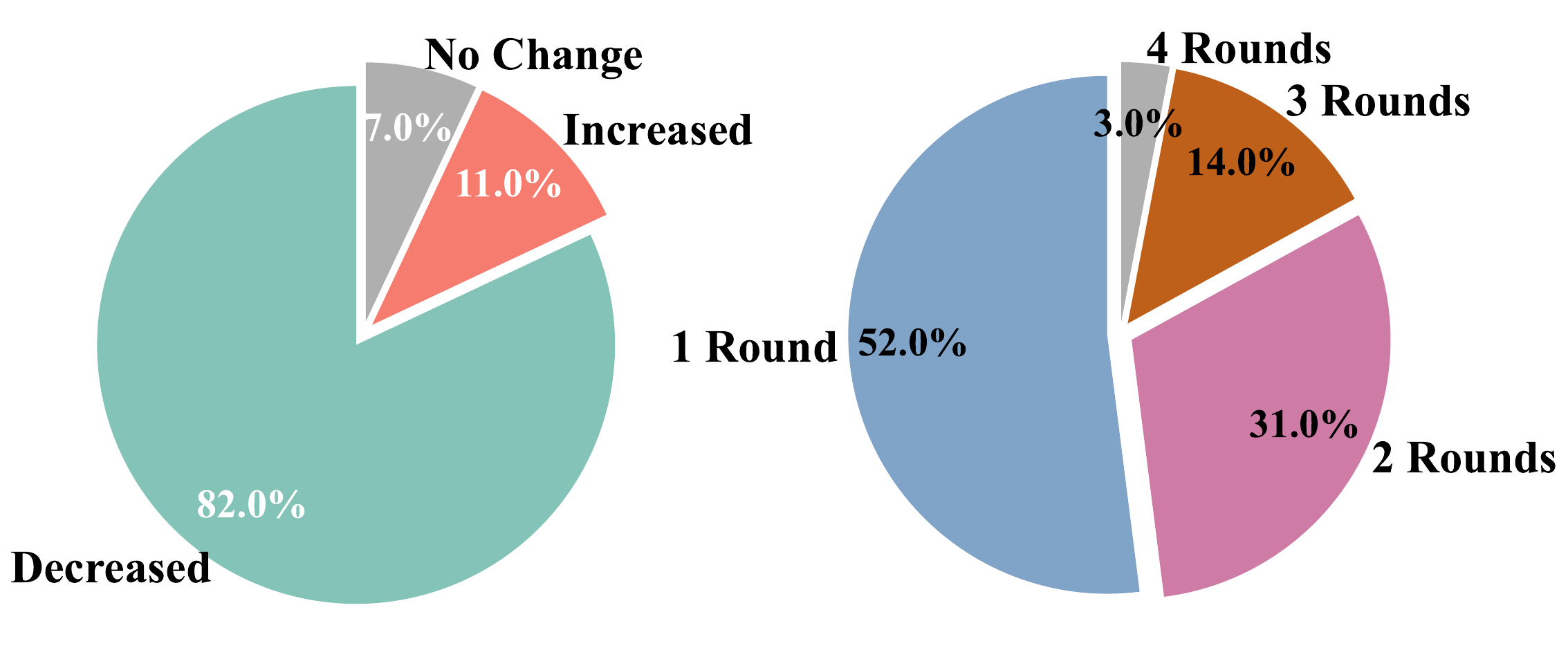}
  \caption{
  (Left) Distribution of Gas optimization effects of real-world contracts.
  (Right) Distribution of Gas optimization loop of real-world contracts.}
  \label{fig:real-gas-effect}
\end{figure}

\begin{figure*}[t]
  \centering
  \includegraphics[width=\linewidth]{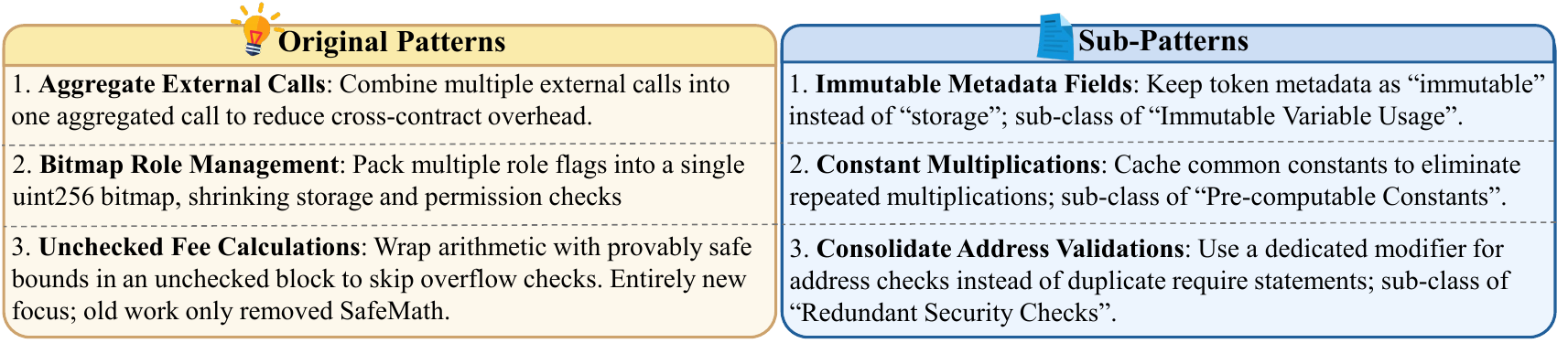}
  \caption{Examples of new patterns discovered by \GasAgent, including both original patterns and sub-patterns.}
  \label{fig:new-pattern}
\end{figure*}

\subsection{Datasets}

We use two datasets to evaluate \GasAgent in different scenarios.

\mypara{Real-world Contracts}
We randomly sample 100 Solidity contracts verified on Etherscan, compiled with version 0.8.20, and deployed after June 2025, to represent diverse real-world deployment styles without any functional category restrictions.

\mypara{LLM-generated Contracts}
Given that the majority of real-world Ethereum smart contracts are associated with the DeFi domain~\cite{wang2025smartcontractsrealworld}, we employ LLMs to generate representative smart contracts for various DeFi protocols.
Specifically, we select the 10 most prevalent DeFi categories based on the number of protocols counted from DefiLlama~\cite{defillama}.
We first use \texttt{GPT-4o} to generate ten core features per category using the corresponding DeFi category description.
Manual inspection confirms that these ten features are sufficient to capture the typical design space of DeFi contracts.
These features are organized into two levels: \textit{Fundamental} and \textit{Advanced}, where the fundamental features simply include all core functionalities and the advanced features incorporate more complex and extended functionalities.
The detailed prompts and configuration for the features generation can be found in our repository.

To ensure that generated contracts are high-quality, we define a contract as \textbf{usable} if it (i) compiles successfully and (ii) passes all fundamental test cases, and for advanced contracts, the additional advanced test cases are applied.
To support this definition, we manually construct 10 test cases for each level in each category.
These test cases are written by two PhD students specializing in blockchain and DeFi, and are carefully designed to match the intended feature semantics while respecting feature dependencies.

We then generate two levels of smart contracts for each category: the fundamental contracts implement only fundamental features, and the advanced contracts include both types of features.
Using the defined features and test suites, we prompt five representative LLMs for smart contract code generation, i.e., \texttt{GPT-4o}~\cite{gpt4o}, \texttt{Llama-4-Maverick-17B-128E-Instruct}~\cite{llama}, \texttt{Gemini-2.5-Flash}~\cite{gemini}, \texttt{DeepSeek-R1-0528}~\cite{deepseek}, and \texttt{Qwen3-235B-A22B}~\cite{qwen}.
For each model, we repeatedly generate both fundamental and advanced contracts using consistent configurations until 5 usable contracts are obtained for each level in each category.

\subsection{RQ1 - Effectiveness}

\Cref{fig:real-gas-distribution} presents the Gas optimization ratios for 100 real-world smart contracts that have been deployed on-chain and successfully verified on Etherscan.
Overall, \GasAgent achieves an average Gas cost reduction of 9.97\%.
Most contracts gain moderate savings in the 5–20\% range, while the top case reaches over 30\%.
A few contracts yield slight negative ratios where attempted changes would increase Gas usage instead.
In this case, \GasAgent falls back to the original smart contract without applying the unhelpful edits.
~\Cref{fig:real-gas-effect} (Left) breaks down these results: 82\% of contracts deliver actual Gas savings, 7\% remain unchanged due to already efficient code, and 11\% would see increased costs if changes were blindly applied.
~\Cref{fig:real-gas-effect} (Right) shows how many rounds were required before the Manager terminates the loop.
52\% of contracts were completed in a single cycle by applying only existing patterns via the Seeker, while 48\% needed at least one round of novel pattern discovery by the Innovator.
Up to four valid new patterns were discovered in some cases, demonstrating \textit{GasAgent}’s capacity to improve contracts beyond existing Gas waste patterns.

\mypara{New Pattern Analysis}
In our evaluation of 100 real-world smart contracts, the Innovator automatically proposes 68 new gas waste patterns.\footnote{Although these new patterns have been empirically validated, we do not directly add them to the Gas waste pattern library.
Instead, they are placed into the Verified New Pattern pool, where human review is optional, allowing for either further manual inspection or direct inclusion as needed.}
During our initial inspection, we observed that some of the newly proposed patterns are not entirely novel, but rather refinements of known patterns, reframed into more specific and actionable forms that may be easily overlooked by humans.
Therefore, to distinguish between refinement patterns and newly discovered patterns, we manually categorize the 68 newly identified patterns into two types: \textit{Sub-patterns} and \textit{Original patterns}, with 30 and 38 patterns in each type, respectively.
Specifically, a sub-pattern represents a refined subclass of an optimization pattern previously reported in the literature, while an original pattern does not belong to any such subclass.
We show some examples of new patterns discovered by \GasAgent for both original patterns and sub-patterns in~\Cref{fig:new-pattern}.
To further illustrate the distinction, we provide detailed examples of one representative original pattern, ``Bitmap Role Management,'' and one representative sub-pattern, ``Immutable Metadata Fields,'' as shown in Listings~\ref{lst:bitmap} and~\ref{lst:immutable}, respectively.

\begin{lstlisting}[language=Solidity, caption={Bitmap Role Management}, label={lst:bitmap}]
// Bad Example
mapping(address => bool) public isAdmin;
mapping(address => bool) public isMinter;

// Gas-Efficient Example
uint256 constant ADMIN  = 1 << 0;
uint256 constant MINTER = 1 << 1;
mapping(address => uint256) private _roles;
\end{lstlisting}

The ``Bitmap Role Management'' is neither included in nor related to our initial Gas Waste Pattern Library.
\GasAgent proposes this pattern to utilize bit operations for compressing the storage structure; specifically, it uses a single ``mapping'' to store all roles associated with an address.
Compared to a gas-inefficient implementation, this approach reduces the storage footprint by one slot per address.
Additionally, the number of getter functions automatically generated by the EVM is reduced by one, resulting in a smaller bytecode size and a contract deployment gas reduction of 96,516.
Besides, by using a \texttt{uint256} bitmap to share a single storage slot, the operations gas cost where a single address corresponds to multiple roles is reduced during execution.

However, ``Immutable Metadata Fields'' is a sub-pattern of ``Immutable Variable Usage'' (which states that any value that can be determined at deployment time and is read-only during execution should be declared as \texttt{immutable} to eliminate costly \texttt{SSTORE}/\texttt{SLOAD} operations) in the initial Gas Waste Pattern Library.
This pattern refines the general concept of immutable variables by specifying concrete categories of metadata that benefit from this optimization and providing actionable implementation strategies.
For metadata fields, declaring them as \texttt{immutable} avoids occupying persistent storage slots.
Instead, their values are embedded into the runtime bytecode as constants during deployment.
As a result, the constructor does not need to perform expensive \texttt{SSTORE} operations, which significantly reduces deployment gas costs.
With this pattern, the code in Listing~\ref{lst:immutable} saves 36,084 gas during deployment.
Furthermore, it also reduces execution-time gas consumption.
Accessing an \texttt{immutable} variable at runtime is reduced to a low-cost constant push operation (e.g., \texttt{PUSH32}), whereas accessing a regular storage variable requires an \texttt{SLOAD}.
Therefore, using \texttt{immutable} for values accessed frequently—such as in high-frequency getters, loops, or per-transaction computations—can lead to substantial gas savings.
The above examples demonstrate the capability of \GasAgent to discover new original patterns, which would typically require extensive experience from smart contract developers and time-consuming manual collection, as well as its ability to identify sub-patterns that refine existing patterns and make them more practical for smart contract development.

\begin{lstlisting}[language=Solidity, caption={Immutable Metadata Fields}, label={lst:immutable}]
// Bad Example
uint256 public chainId;
uint256 public launchTimestamp;

// Gas-Efficient Example
uint256 public immutable chainId;
uint256 public immutable launchTimestamp;

constructor(uint256 _chainId, uint256 _launchTimestamp) {
    chainId = _chainId;
    launchTimestamp = _launchTimestamp;
}
\end{lstlisting}

\begin{table}[t]
\caption{Distribution of the 68 newly identified Gas-saving patterns discovered by \GasAgent, categorized by their optimization method.}
\label{tab:gasagent-stats}
\centering
\begin{tabular}{l|c|c|c}
\toprule
\textbf{Category} &
\textbf{Original} &
\textbf{Sub} &
\textbf{Total} \\
\midrule
Batch \& Consolidate              & 10 & 18 & 28 \\
Mapping, Struct Data             & 16 &  6 & 22 \\
Bitwise, Packing \& Unchecked    & 10 &  0 & 10 \\
Misc.\ Safe Compute \& Storage    &  2 &  6 &  8 \\
\midrule
\textbf{Overall}                  & 38 & 30  & 68 \\
\bottomrule
\end{tabular}
\end{table}

\begin{figure}[t]
  \centering
  \includegraphics[width=\linewidth]{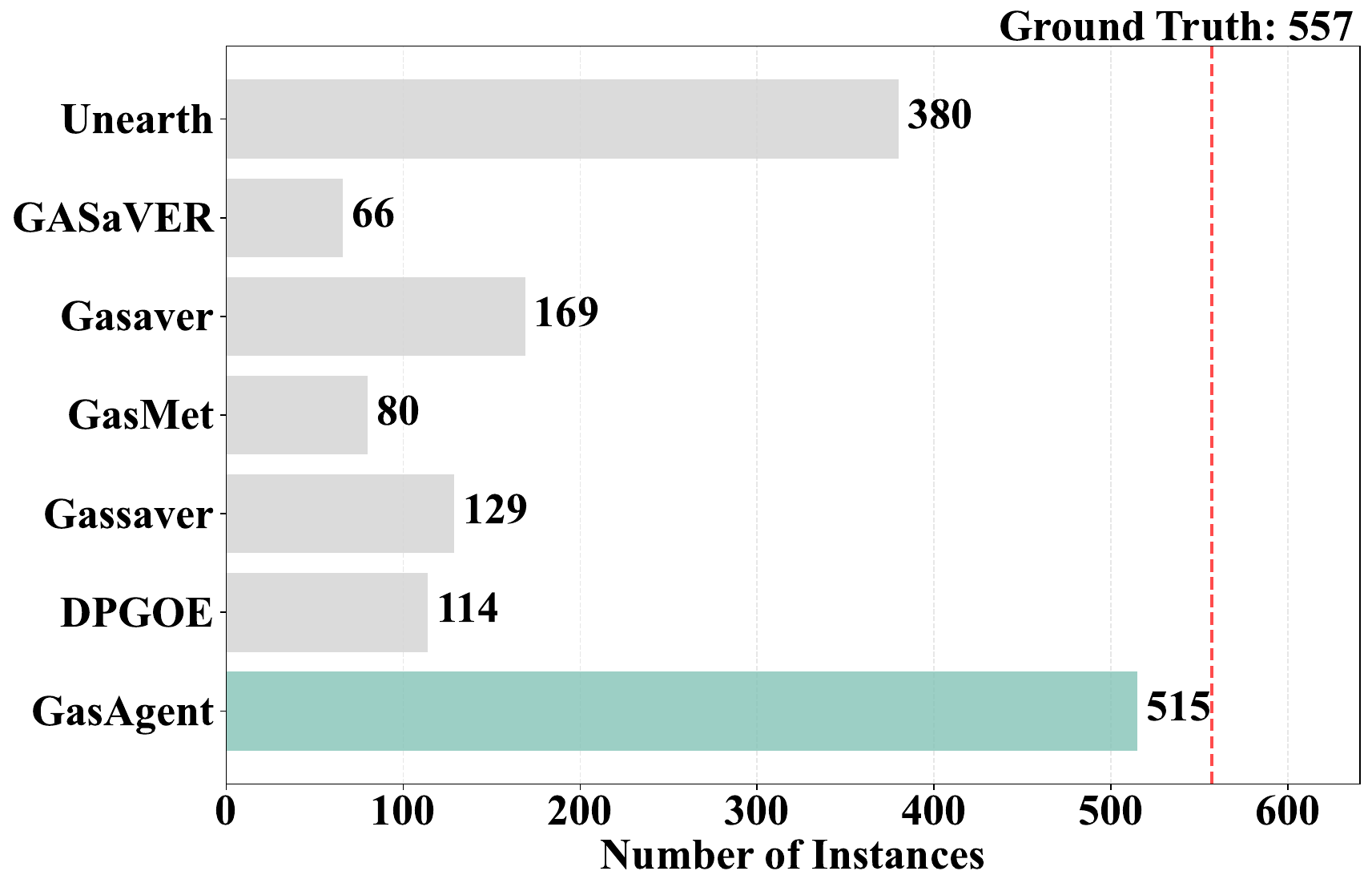}
  \caption{
  Result of pattern incorporation test, which contains: Unearth~\cite{jiang2024unearthing}, GASaVER~\cite{kaur2022gas}, Gasaver~\cite{zhao2022gasaver}, GasMet~\cite{di2022profiling}, Gassaver~\cite{nguyen2022gassaver}, and DPGOE~\cite{marchesi2020design}. 
  \GasAgent retrieves 515 out of 557 instances by integrating prior patterns.
  }
  \label{fig:baseline-compatibility}
\end{figure}

Such fine-grained variants show how small implementation details can yield Gas savings that general guidelines might miss, demonstrating that the new patterns discovered by \GasAgent complement expert knowledge.

Besides that, as shown in~\Cref{tab:gasagent-stats}, we also categorized these patterns into groups based on their optimization methods.
As can be seen, the 68 patterns identified by \GasAgent almost cover all key aspects of smart contract gas optimization: ranging from high-level transaction flow optimizations such as batching and consolidation, to improvements in storage layout involving mappings and struct data, and further down to low-level optimizations in bitwise operations, data packing, and arithmetic bounds.
The dominance of the categories ``Batch \& Consolidate'' and ``Mapping, Struct Data'' reflects the fact that gas-intensive operations in real-world contracts often occur in batch processing and data organization.
Real-world contract profiling pinpoints batch workflows and data layout as the chief gas sinks—blind spots most developers miss.
Details of these 68 patterns mentioned above are all listed in this repository.

\begin{tcolorbox}[colback=black!10, colframe=black!50, boxrule=0.5pt, arc=3pt, left=6pt, right=6pt, top=4pt, bottom=4pt, before skip=10pt, after skip=10pt]

\textbf{Takeaway for RQ1:} \GasAgent demonstrates strong effectiveness in reducing Gas costs.
It successfully optimized 82\% of 100 real-world contracts, achieving an average Gas saving of 9.97\%.
In addition, it discovered 68 new patterns that were validated in the real-world contracts optimization process, confirming its capability to contribute meaningful Gas-saving strategies in practice.

\end{tcolorbox}

\subsection{RQ2 - Pattern Incorporation}

To evaluate whether \GasAgent incorporates and reuses existing Gas waste patterns, we construct a test using a library of patterns collected from prior work.
We survey six representative studies~\cite{jiang2024unearthing,kaur2022gas,zhao2022gasaver,di2022profiling,nguyen2022gassaver,marchesi2020design} and extract 24 distinct Gas waste patterns that they propose.
Each pattern is implemented as a standalone Python detection tool that accepts Solidity source code as input and outputs matching opportunities along with suggested transformations.
All tools are integrated into the Gas Waste Pattern Library of \GasAgent and can be retrieved and activated by the Seeker.

We use the parameter shown in~\Cref{sec:setting} and run 24 tools exhaustively over 100 real-world contracts to build the ground truth: if a tool finds a valid match, we record the corresponding pattern as present.
Next, we run the Seeker’s Natural Language-based and code-similarity retrieval pipeline on the same contracts to check whether it correctly activates the relevant tools.
As shown in~\Cref{fig:baseline-compatibility}, Seeker successfully retrieves 515 out of 557 ground-truth instances, achieving a recall of 92.5\%.
This result demonstrates that patterns proposed across different prior studies, while each covering only a subset of the instances, can be effectively consolidated and reused within \textit{GasAgent}’s unified framework.

It is important to note that this recall is not an upper bound, but rather a threshold-controlled result.
If we set the retrieval threshold to zero, which can let \GasAgent achieve 100\% recall, at the cost of increased computation.
With the current threshold, Seeker reduces the number of tool calls from 2,400 to 1,722—a 28.25\% reduction—while still preserving high coverage.
Such efficiency gains are particularly valuable as the pattern library continues to expand with more specialized or resource-intensive tools.

\begin{tcolorbox}[colback=black!10, colframe=black!50, boxrule=0.5pt, arc=3pt, left=6pt, right=6pt, top=4pt, bottom=4pt, before skip=10pt, after skip=10pt]

\textbf{Takeaway for RQ2:} \GasAgent effectively incorporates the existing Gas waste pattern.
It recalled 92.5\% of all existing pattern instances across 100 real-world contracts while reducing detection calls by 28.2\%.
This demonstrates that Seeker can efficiently utilize prior patterns through dual retrieval, achieving high coverage with fewer detection calls.

\end{tcolorbox}

\subsection{RQ3 - Design Rationality (Ablation Study)}

To verify that the Seeker and the Innovator of \GasAgent contribute meaningfully to overall Gas optimization, we conduct an ablation study comparing three stripped-down variants:  
(1) \textbf{Direct LLM}: directly prompting the same \texttt{GPT-4o} model (as described in~\Cref{sec:setting}) to refactor the contract without any explicit pattern retrieval or multi-agent orchestration;
(2) \textbf{Without Innovator}: running \GasAgent without the Innovator, covering known patterns but disabling new pattern discovery;  
(3) \textbf{Without Seeker}: running \GasAgent without the Seeker, discovering new patterns but ignoring the curated pattern library.

\begin{figure}[b]
  \centering
  \includegraphics[width=\linewidth]{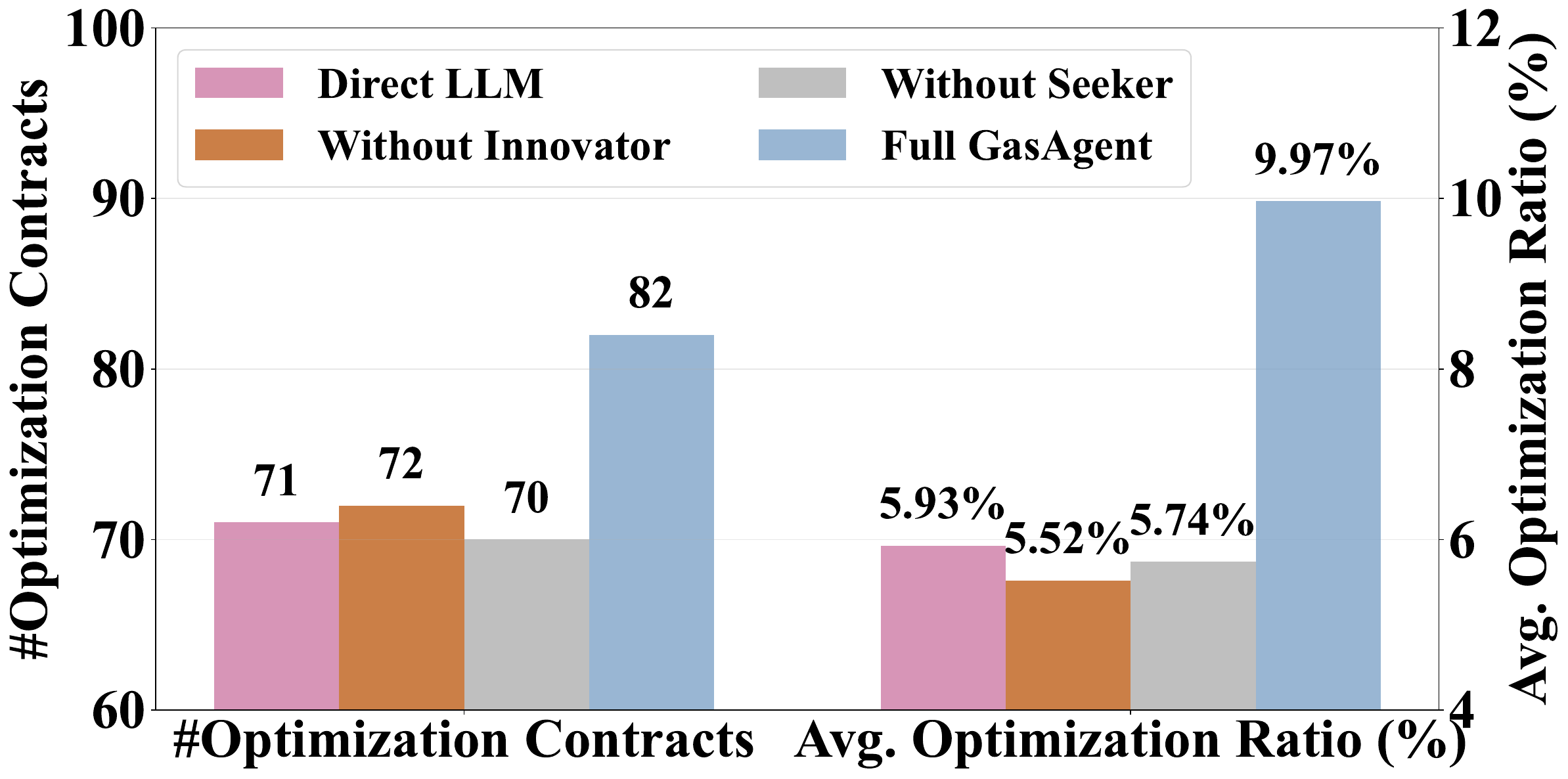}
  \caption{
  Result of ablation study: Full \GasAgent achieves superior effectiveness compared to partial \GasAgent or LLM-only approaches.
  }
  \label{fig:ablation-study}
\end{figure}

~\Cref{fig:ablation-study} shows the results of 100 real-world contracts.
The direct LLM can only optimize 71 out of 100 contracts, with an average Gas saving of 5.93\%.
Disabling the Innovator module while keeping the Seeker results in 72 optimized contracts (5.52\% average saving).
Conversely, removing the Seeker but keeping the Innovator covers 70 contracts (5.74\% average saving).
In contrast, the full \GasAgent system achieves the highest coverage, optimizing 82 contracts, with an average saving of 9.97\%.

These results confirm that both the Seeker (for existing pattern reuse) and the Innovator (for discovering new patterns) are essential.
Interestingly, using either alone does not outperform direct LLM rewriting, which we attribute to the fact that the Seeker and the Innovator are designed to work collaboratively.
When run in isolation, their prompts are more narrowly scoped—either for tool invocation or pattern innovation—whereas a direct LLM rewriting prompt is more general-purpose and covers a wider search space.
This highlights that \textit{GasAgent}’s strength lies in combining dedicated modules in a loop, not in using any single one in isolation.

\begin{tcolorbox}[colback=black!10, colframe=black!50, boxrule=0.5pt, arc=3pt, left=6pt, right=6pt, top=4pt, bottom=4pt, before skip=10pt, after skip=10pt]

\textbf{Takeaway for RQ3:} 
The ablation study confirms that \textit{GasAgent}’s collaborative design is crucial: while a single direct LLM rewrite or partial agent use yields modest savings, combining the Seeker and Innovator within the full framework consistently achieves higher Gas cost reduction in both scope and depth.

\end{tcolorbox}

\begin{table*}[htbp]
\centering
\caption{Optimization results of \GasAgent on smart contracts generated by five representative LLMs under two task difficulty levels. 
Each subtable reports the average original Gas cost, the optimized Gas cost, the average saving percentage, and the number of contracts successfully optimized out of 50.
}
\label{tab:llm-gas-comparison}
\resizebox{0.98\linewidth}{!}{
\begin{tabular}{l|cccc|cccc}
\toprule
\multirow{2}{*}{\textbf{LLM}} & \multicolumn{4}{c|}{\textbf{Fundamental Contracts}} & \multicolumn{4}{c}{\textbf{Advanced Contracts}} \\
\cmidrule{2-5} \cmidrule{6-9}
 & \textbf{Original Gas} & \textbf{Optimized Gas} & \textbf{Saving (\%)} & \textbf{Count} 
 & \textbf{Original Gas} & \textbf{Optimized Gas} & \textbf{Saving (\%)} & \textbf{Count} \\
\midrule
\textbf{Qwen3}         & 847,985 & 803,794 & 7.27 & 43/50 & 1,597,971 & 1,555,050 & 5.45 & 33/50 \\
\textbf{Llama-4}   & 609,023 & 517,280 & 13.93 & 50/50 & 954,106   & 878,739   & 8.74 & 46/50 \\
\textbf{DeepSeek-R1}        & 824,826 & 795,372 & 5.77 & 39/50 & 1,526,594 & 1,516,740 & 5.46 & 30/50 \\
\textbf{Gemini-2.5}   & 1,121,518 & 1,030,543 & 9.24 & 40/50 & 2,307,843 & 2,233,976 & 4.79 & 36/50 \\
\textbf{GPT-4o}& 596,964 & 550,286 & 9.89 & 41/50 & 809,951   & 755,826   & 8.36 & 41/50 \\
\bottomrule
\end{tabular}
}
\end{table*}

\subsection{RQ4 - Broader Usability}

\begin{figure}[b]
  \centering
  \includegraphics[width=\linewidth]{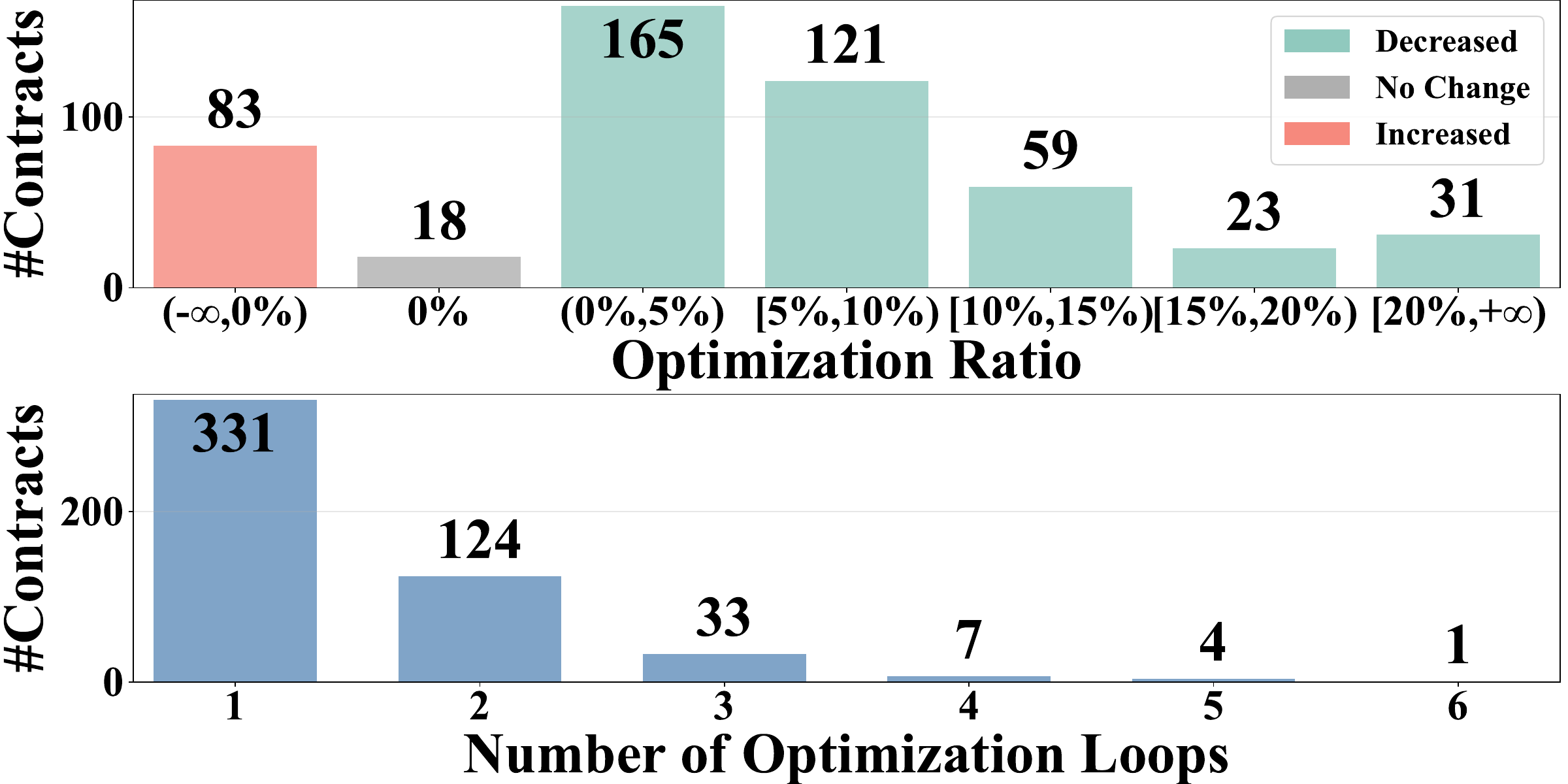}
  \caption{
  (Top) Distribution of LLM-generated contracts by Gas optimization ratio.
  (Bottom) Distribution of LLM-generated contracts by number of optimization loops.
  }
  \label{fig:llm-gas-distribution}
\end{figure}

To assess \textit{GasAgent}’s broader usability, we evaluate its effectiveness as an automated optimization layer within LLM-assisted smart contract development workflows.
~\Cref{fig:llm-gas-distribution} presents the impact of \GasAgent when applied to contracts generated by LLMs.
Compared to real-world contracts in ~\Cref{fig:real-gas-distribution} and ~\Cref{fig:real-gas-effect}, the distribution is broadly similar but shows a slightly higher concentration in the small-to-moderate saving ranges.
In total, 79.8\% of the LLM-generated contracts are successfully optimized, with 57.2\% achieving savings in the (0\%–10\%] range.
The remaining 20.2\% either show no measurable gain or a negligible Gas increase; for these cases, \GasAgent safely retains the original version without applying any changes.
66.2\% of contracts terminate their optimization after just one pass—slightly higher than the 52\% for real-world contracts—while several more complex examples require up to five or six iterations to find and verify valid new patterns.

To complement this, ~\Cref{tab:llm-gas-comparison} details \textit{GasAgent}’s impact across five representative LLMs under both fundamental and advanced task conditions.
The results show that while all tested LLMs leave some redundant Gas usage that \GasAgent can automatically optimize, the amount of optimization ratio(saving) and the success rate vary noticeably across models.
For example, for fundamental contracts, \texttt{Llama-4} achieves the highest average savings at 13.93\% and fully optimizes all 50 test contracts (50/50), whereas \texttt{DeepSeek-R1} only has 5.77\% savings with 39 out of 50 successfully optimized.
This indicates that different LLMs generate Solidity code with distinct structural tendencies—some produce more straightforward code patterns that align closely with known or discoverable Gas waste, while others tend to generate more compact or unconventional structures that are harder to match with existing patterns.

The table also reveals that as task complexity increases, both the average savings and the proportion of successfully optimized contracts drop consistently across all tested models.
For instance, \texttt{Qwen3-235B}’s average saving falls from 7.27\% to 5.45\%, with the success count dropping from 43/50 to 33/50.
Similarly, \texttt{Llama-4} drops from 13.93\% (50/50) to 8.74\% with a slight decrease to 46/50, and \texttt{Gemini-2.5} decreases more sharply from 9.24\% (40/50) to just 4.79\% with only 36 contracts optimized out of 50.
We suspect that higher-complexity contracts involve deeper nesting, more dynamic control flows, or cross-function state dependencies that reduce the effectiveness of pattern-based static matching and make new pattern discovery and automatic validation more challenging.
In these cases, parts of the residual inefficiency may remain hidden from \textit{GasAgent}’s current detection pipeline, especially when redundant operations are entangled with functional logic that must not be altered.
Verifying this hypothesis more rigorously—such as by combining dynamic execution traces or deeper semantic flow analysis—remains an important direction for improving \textit{GasAgent}’s coverage on advanced contract code.

Despite these limitations, \GasAgent still achieves non-trivial savings even for complex contracts, demonstrating its robustness as an automated top-up layer that provides developers with Gas improvements without extra manual effort.
The visible differences among LLMs further show that \GasAgent can serve as a practical diagnostic tool, revealing which kind of LLMs tend to leave more hidden redundancies under the same conditions.

\begin{tcolorbox}[colback=black!10, colframe=black!50, boxrule=0.5pt, arc=3pt, left=6pt, right=6pt, top=4pt, bottom=4pt, before skip=10pt, after skip=10pt]

\textbf{Takeaway for RQ4:} Overall, \GasAgent optimizes nearly 80\% of LLM-generated contracts with average savings ranging from 4.79\% to 13.93\% depending on the model and task complexity.
This confirms its practical usability as a reliable automated optimization layer that not only removes residual Gas waste left by LLMs but also reveals meaningful structural differences across generation pipelines.

\end{tcolorbox}

\section{Related Work}

Gas optimization in Solidity smart contracts remains a key challenge due to the platform’s cost model and low-level execution semantics.
Existing techniques are typically classified into \textit{compiler-level optimization}, \textit{code-smell-based rewriting}, and \textit{super-optimization}.
At the compiler level, solc applies default peephole optimizations and optional Yul-level optimizations such as dead assignment elimination and expression folding~\cite{solc}.
Since version 0.8.0, Solidity has introduced automatic overflow checks and the \texttt{unchecked} block to reduce gas~\cite{FN}.
However, many inefficiencies—such as redundant storage access, missing \texttt{calldata} annotations, and suboptimal visibility—remain beyond the reach of compiler-level strategies~\cite{he2024save}.

A significant portion of prior research~\cite{chen2017under,chen2020gaschecker,zhao2022gasaver,nguyen2022gassaver,kaur2022gas,di2022profiling,marchesi2020design,chen2018towards} focuses on \textit{code-smell-based optimization}, which relies on expert-defined gas-inefficient patterns and static analysis to detect and refactor suboptimal code.
GASPER~\cite{chen2017under} identifies dead code and redundant loops.
Its successors, GasReducer~\cite{chen2018towards} and GasChecker~\cite{chen2020gaschecker}, expand pattern coverage and execution scalability, though neither is publicly available.
GasSaver~\cite{kaur2022gas} implements a small set of rule-based checkers targeting Solidity-specific inefficiencies such as missing \texttt{calldata}, improper visibility, and unnecessary array reads.
While useful, these tools are constrained by their reliance on hand-crafted rules and offer limited coverage of modern gas usage patterns, lacking adaptability to diverse contract structures or emerging inefficiencies.
Recently, Jiang et al.~\cite{jiang2024unearthing} explored using a single LLM to detect new gas waste code patterns in Solidity.
While the approach is promising, it suffers from hallucinations and redundant suggestions, making it difficult to ensure the novelty and correctness of the identified patterns.
A third line of work adopts \textit{super-optimization}, which formulates gas optimization as a formal search problem.
GASOL~\cite{albert2020gasol} and loop-based optimization~\cite{nelaturu2021smart} use symbolic reasoning or SMT solvers to synthesize more efficient code.
While effective in specific cases, these methods are often computationally expensive and hard to scale to real-world contracts.

\section{Conculsion}
We present \textbf{GasAgent}, the first multi-agent framework for end-to-end Gas optimization in smart contracts.
By combining compatibility with existing Gas-saving patterns and automated discovery and validation of new ones, GasAgent addresses key limitations of both manual auditing and LLM-only approaches.
It consists of four collaborative agents that identify, evaluate, and apply optimizations in a closed-loop workflow.
Experiments on 100 real-world contracts show that GasAgent improves 82\% of them with an average Gas reduction of 9.97\%, and generalizes well to LLM-generated contracts across diverse categories.
We hope this work contributes to the broader effort of building automated tooling for efficient and intelligent smart contract development.

\bibliographystyle{unsrtnat}
\bibliography{ref}

\end{document}